\pdfoutput=1
\documentclass[11pt,table]{article}

\usepackage[preprint]{acl}

\usepackage{times}
\usepackage{latexsym}

\usepackage[T1]{fontenc}
\usepackage[utf8]{inputenc}

\usepackage{microtype}

\usepackage{inconsolata}

\usepackage{graphicx}

\usepackage{ulem}
\usepackage{array} % for more column types
\usepackage{multirow}
\usepackage[T1]{fontenc}
\usepackage{lmodern}
\usepackage{verbatim}
\usepackage[english]{babel}
\usepackage{listings}  % write xml code and other source codes
\usepackage{url}
\usepackage{amsmath}
\usepackage{paralist}  % Import compactenum class
\usepackage{color}
\usepackage{theorem}
\usepackage{amsfonts}
\usepackage{amssymb}
\usepackage{amsmath}
\usepackage{bm}
\usepackage{hyperref}

\title{LengClaro2023: A Dataset of  Administrative Texts in Spanish with Plain Language adaptations}

\author{Belén Agüera-Marco \\
  University of the Basque Country UPV/EHU \\
  \texttt{baguera001@ikasle.ehu.eus} \\\And
  Itziar Gonzalez-Dios \\
  HiTZ Center - Ixa \\
 University of the Basque Country UPV/EHU\\
  \texttt{itziar.gonzalezd@ehu.eus} \\}

\begin{document}
\maketitle
\begin{abstract}

In this work, we present LengClaro2023, a dataset of legal-administrative texts in Spanish. Based on the most frequently used procedures from the Spanish Social Security website, we have created for each text two simplified equivalents. The first version follows the recommendations provided by arText claro. The second version incorporates additional recommendations from plain language guidelines to explore further potential improvements in the system. The linguistic resource created in this work can be used for evaluating automatic text simplification (ATS) systems in Spanish.
\end{abstract}

\section{Introduction}

In this report, we present a part of the master thesis written by Belén Agüera Marco in order to obtain the B.S. Language Analysis and Processing  at the University of the Basque Country (UPV/EHU), supervised by  Itziar Gonzalez-Dios.

In today's information-driven society, access to legal-administrative texts is crucial for citizens to exercise their rights and fulfill their obligations. However, the complexity of such texts often creates barriers. The objective of this Master's Thesis is to contribute to the improvement of the accessibility of legal-administrative texts in Spanish for citizens. This will be done by constructing and analysing LengClaro2023\footnote{\href{https://github.com/baguera001/LengClaro2023}{https://github.com/baguera001/LengClaro2023}}, a dataset that compiles texts available on the official website of the Spanish Social Security\footnote{\href{www.seg-social.es}{www.seg-social.es}} in their original format as well as two simplified versions of each text shared with CC BY-NC-SA 4.0 license. 

Text simplification is the process of reducing the linguistic complexity of a text while still retaining the original information content and meaning \citep{Siddharthan2014ASO}. This involves modifying the content and structure of the text in order to improve its understandability and make it easier to read \citep{alva-manchego-etal-2020-data}. 
In this work, we have implemented these modifications to ensure that the texts adhere to the principles of Plain Language. As defined by the International Plain Language Federation: 
\begin{quotation}
  A communication is in plain language if its wording, structure, and design are so clear that the intended readers can easily find what they need, understand what they find, and use that information.
\end{quotation}

Although Plain Language share with Easy-to-Read (E2R)\footnote{\href{https://www.inclusion-europe.eu/easy-to-read/}{www.inclusion-europe.eu/easy-to-read/}} the common goal of making information more comprehensible through textual simplification, it is important to distinguish between the two approaches. E2R is specifically designed to meet the needs of individuals with comprehension difficulties, such as those with intellectual disabilities, cerebral palsy, or autism spectrum disorder, as highlighted by the organization Plena Inclusión\footnote{\href{https://www.plenainclusion.org/noticias/cual-es-la-diferencia-entre-lectura-facil-y-lenguaje-claro/}{www.plenainclusion.org/noticias/cual-es-la-diferencia-entre-lectura-facil-y-lenguaje-claro/} [Accessed: 20th August 2024]}. In contrast, Plain language is intended for the general public, which aligns with our goal of improving the clarity of administrative communications for the citizenry.

We selected seven key pages of information and created two simplified versions of each. The first set of simplified texts was developed based on the recommendations provided by arText claro, a tool designed to assist administrative staff in communicating more clearly with the public. A qualitative analysis was conducted to assess the effectiveness of these recommendations, considering the specific characteristics of the texts published on the website and the ease of implementation from the perspective of a user of the tool. In order to create the second set of simplified versions, we conducted a bibliographical review of plain language recommendations for legal-administrative texts in the recent years. Based on this review, and incorporating insights gained from the analysis of arText claro, we created a new set of recommendations. These were applied without the assistance—yet also without the limitations—of automatic detection of problematic linguistic features.

Based on the findings of these analyses, we propose several modifications and new recommendations for arText claro. By enhancing the effectiveness of text simplification tools and making them more responsive to the specific needs of public administrators and citizens, we aim to contribute to the creation of more accessible and inclusive public communication. We also expose the main linguistic challenges encountered during the text simplification process with the objective of bridging the gap between the theoretical principles of plain language and their practical application. 

 In the following sections, we present the dataset compilation (Section \ref{sec:dataset}), the simplifications carried out with the tool Artext Claro (Section \ref{sec:artext-claro}), the simplifications carried out based on Plain Language guidelines (Section \ref{sec:LengClaro-simplification}), the discussion (Section \ref{sec:chap5-disc}) and the conclusion and future work (Section \ref{sec:conc}).

\section{Dataset compilation}
\label{sec:dataset}

In this section, we expose the process undertaken to create the LengClaro2023 dataset. The compilation involved selecting the original legal-administrative texts according to their relevance to the goals of this work
and extracting and structuring them to form a coherent and functional dataset. For the selection of the texts, we explain the motivation and criteria behind the choices of the most suitable texts to the research objectives. Secondly, we detail the necessary steps to extract the texts and how the dataset is organized to ensure the data's consistency. 

\subsection{Selection of texts}
\label{sec:texts-selection}

In order to build a LengClaro2023, we determined that the most beneficial approach would be to focus on texts from an institution that deals with complex processes and has a direct impact on citizens' daily lives. Therefore, we decided to use Spanish Social Security communications, as the decisions and actions that it covers—such as those related to pensions, subsidies, and public health care—are often confusing, concern vulnerable situations, and play a significant role in people's lives. Simplifying the language used by Social Security will help people better understand their rights and duties and facilitate access to essential services and benefits. Additionally, Social Security affects a large population, which may amplify the impact of the results of this work.

Since our objective is to address communications that are particularly beneficial for the general public, we decided to focus on the content that Social Security makes available to citizens on its web portal. Nowadays, when the administration needs to communicate with the general public rather than a specific individual, it primarily does so through these online platforms \citep{Pistola2022}. We felt it was most appropriate to focus on two sections in particular that fulfill the requirement of having content specially useful to citizenship. On the homepage, there is a section %highlighted in blue
titled "A un click" (One click), which includes links to various informational pages. This indicates that the administration considered these contents especially important and wanted to make sure they are easy for citizens to access. Consequently, we prioritized the content within these links. Similarly, we chose the homepage tab "Lo más visitado" (The most visited) to complete our collection, based on the assumption that the pages in this section are the most frequently consulted by the public.

Some of the pages accessible through the links in the mentioned sections did not contain textual information. One featured video content and another provided access to a specific service requiring digital certificate authentication. Given that our focus is on linguistic simplification, we excluded these pages from consideration\footnote{The pages excluded were:
\href{https://w2.seg-social.es/fs/indexframes.html}{Acceso Sistema RED on-line} and
\href{https://ismarina.doopaper.net/publicacion/ismarina/ismarina8002/Video-Esperanza-del-mar--NIPO-WEB--273-17-067-x.html}{Buques asistenciales ISM}.
[Accessed: 23rd October 2023]}.
Additionally, other links in the sections did not provide direct information but rather included multiple links to other pages. Therefore, we made the selection only across those links that led to pages where simplification could enhance accessibility\footnote{The pages excluded because of lack of written content were:
\href{https://www.seg-social.es/wps/portal/wss/internet/InformacionUtil/262/12032/}{Compra y venta de inmuebles},
\href{https://sede.seg-social.gob.es/wps/portal/sede/sede/Ciudadanos/cita previa para pensiones y otras prestaciones/13cita previa para pensiones y otras prestaciones/}{Cita previa para prestaciones y otras gestiones},
\href{https://www.seg-social.es/wps/portal/wss/internet/InformacionUtil/44539}{Trámites y gestiones},
\href{https://www.seg-social.es/wps/portal/wss/internet/OficinaSeguridadSocial}{Direcciones y teléfonos},
\href{https://www.seg-social.es/wps/portal/wss/internet/InformacionUtil/5300/1490}{Acceso al Sistema RED},
\href{https://www.seg-social.es/wps/portal/wss/internet/HerramientasWeb/4c5c5105-04d1-4dfd-8af4-1e2d5c0c1ae3}{Cómo identificarme},
\href{https://www.seg-social.es/wps/portal/wss/internet/Pensionistas}{Pensionistas},
\href{https://www.seg-social.es/wps/portal/wss/internet/InformacionUtil/5300}{Sistema RED} and
\href{https://www.seg-social.es/wps/portal/wss/internet/Trabajadores}{Trabajadores}. [Accessed: 23rd October 2023]}.

In particular, we focused on links to pages that provide information for people who would most benefit from language simplification due to the importance of the procedure involved,  such as those aimed at combating exclusion and vulnerability. By reducing the possibility of errors in the application process, it can also help avoid unnecessary delays or rejections. Furthermore, it reduces the need for advisors. This is especially important for those who cannot afford this additional service and for older people, who become more autonomous if they can manage their own affairs without relying on family or friends.

Below is a brief summary of the content that will form part of our dataset:

\begin{itemize}
    \item \textbf{Document 1. Ingreso Mínimo Vital (Minimum Vital Income).} This page explains a financial benefit designed to guarantee a minimum level of income and prevent the risk of poverty and social exclusion. It outlines the most important aspects, such as requirements, amounts, or incompatibilities.
    
    The formal and convoluted language used in the Minimum Vital Income (MVI) information page was highlighted by a RTVE news item from June 2022 \href{https://www.rtve.es/noticias/20220608/lenguaje-claro-administracion-ciudadanos/2349417.shtml}{\textit{En lucha contra el lenguaje enrevesado de la Administración}}, which reported that most administrative texts in Spain are not clear, specially those that explain how to apply for a grant, scholarship, or subsidy.
    
    \item \textbf{Document 2. Nacimiento y Cuidado de Menor (Birth and child care).} This page exposes information about the service to claim childbirth and childcare allowance.
    
    \item \textbf{Document 3. Pago de deudas con tarjeta (Payment of debts by debit or credit card).} In this page it is offered information about the Social Security service for paying debts by bank card.
    
    \item \textbf{Document 4. Violencia contra la mujer (Violence against women).} This page brings together the measures offered by the Social Security to alleviate the effects suffered by victims of violence against women, such as orphan's pensions or exceptions in their favour in retirement matters.
    
    \item \textbf{Document 5. Acreditación de vivencia (Proof of life).} This page presents the different options available for pensioners living abroad to provide their proof of life. In particular, it details the functioning of VIVESS, a mobile application for this purpose. Pensioners are usually old people, who often struggle with new technologies, so it is very important to make it easy for them to be able to interpret this kind of information.
    
    \item \textbf{Document 6. Certificado integral de prestaciones (Comprehensive Benefit Certificate).} This page provides information on the service that allows you to obtain certificates on pensions received in a single document. 
    
    \item \textbf{Document 7. Tarjeta Sanitaria Europea (European Health Insurance Card).} This page explains the means and conditions for applying for a document to receive health benefits in the European Economic Area.
\end{itemize}

The linguistic characteristics of the texts selected for LengClaro2023 are consistent with classic textual genres of the legal-administrative field. However, it is important to note that these texts also have specific peculiarities due to their status as web content. Our texts are in fact hypertexts. This concept, introduced by \citet{nelson1981literary}, refers to digital texts structured in blocks of information connected by links. This type of organization allows readers to access content in a non-sequential way. The information on the web pages that comprise our dataset is organized into sections with titles and usually includes a clickable table of contents at the beginning, which enables users to quickly jump to the part they are most interested in. This expands the possibilities for reading, and it cannot be assumed in which order the reader will approach the text.

\subsection{Extraction of texts and structure of the dataset}
\label{sec:texts-extraction}
We extracted the content of the mentioned web pages using the Linux command "wget". When executed, this command generates an HTML file containing the code of the specified URL. 

For each of our documents, we carried out two simplification processes. The first process followed the recommendations proposed by the arText claro system, which are explained in detail in Section \ref{sec:artext-claro}. The second process involved simplifying the original texts without the restrictions of the automatic detection of problematic elements, following the guidelines outlined in Section \ref{sec:LengClaro-simplification}.
These simplifications were also brought into an HTML format.

The files were then renamed to reflect their corresponding document number, as exposed in Section \ref{sec:texts-selection}, followed by an indication of the version condition. Since each document has three versions, we have three version tags: "original" for the files extracted from the Social Security website, "artext" for the files generated on the basis of arText claro recommendations, and "lengclaro", for simplifications made without the writer assistant. The naming convention follows the format:
\begin{quote}
    [document number]\_[version].html.
\end{quote}

For instance, the file containing the text about birth and child care extracted from the Social Security website was named \textit{2\_original.html}.

Consequently, the structure of the dataset built in this work consists of sets of HTML trios: original, artext, and LengClaro files. The original texts from the Spanish Social Security maintain their original licenses, while our simplified adaptations are shared under the CC BY-NC-SA 4.0 license.

\section{Simplifications with arText claro}
\label{sec:artext-claro}

For the first simplified version of the documents, we relied on arText claro's Plain Language Recommendations Module. It was not designed for hypertexts, which may affect the suitability of some recommendations, but we used it as a good approximation since it deals specifically with administrative language. The Spelling and Formatting Module is not particularly relevant for linguistic simplification, and the Text Structure and Content Module was not designed to advise on texts for a web portal.

The process consisted of the following steps:
\begin{enumerate}
\item We extracted the textual content, including extra-linguistic features such as headings, subheadings, lists, and so on, from the web pages and entered it into the text editor. This was necessary because the system's text editor does not handle text in HTML format correctly. Additionally, arText claro does not perform well with very long texts, such as Document 1. Therefore, the longer texts had to be divided into sections for the system to process them effectively.
\item We clicked on \textit{Revisar el texto} (Revise the text) to generate recommendations.
\item We reviewed each recommendation and made the necessary amendments to the text.
\item We clicked again on \textit{Revisar el texto} to check if the new text met the expectations. If not, we went back to step 3. 
\item After completing the simplifications, we adapted the simplified texts back into HTML format to maintain consistency within the dataset and to preserve web features.
\end{enumerate}

During the simplification process, the system occasionally failed to run the review and remained stuck in processing, regardless of the text length. Additionally, it sometimes did not return any recommendations at all, or returned a recommendation without highlighting any part of the text to apply it. These technical drawbacks may discourage users from relying on the tool.

Below, we present the recommendations provided by arText claro, organized into three levels of language: discourse, morphosyntactic, and lexical \citep{da2022redactor}\footnote{The full texts of the recommendations provided by arText claro can be found in Annex 1 of this article.}. We also present the insights gained from simplifying the aforementioned documents. 

\paragraph{a. Recommendations for the Discourse Level}\textcolor{white}{.}
\label{par:artext-a}
\subparagraph{a1. Revision of Sentence-Paragraphs.}
\label{par:artext-a1}

arText claro recommends that each paragraph should include at least two sentences. For example, the following paragraphs were marked for revision: 
\begin{quote}
    La Seguridad Social limita en todo caso el acceso y utilización de los datos personales.
    
    Además, las víctimas de violencia contra la mujer pueden solicitar, a través de nuestras Oficinas de la Seguridad Social, que el acceso a sus datos queden especialmente limitado y controlado.

\end{quote}

The solution for the problem involved combining both sentence-paragraph in a single paragraph consisting of three sentences:
\begin{quote}
    La Seguridad Social limita en todo caso que se acceda y utilicen los datos personales. Además, las víctimas de violencia contra la mujer pueden solicitar que el acceso a sus datos quede especialmente limitado y controlado. Esto pueden solicitarlo a través de nuestras Oficinas de la Seguridad Social.
\end{quote}

However, the system does not always accurately identify sentence-paragraphs. In some cases, it fails to detect them for user revision. More often, it incorrectly labels fragments of text as "sentence-paragraph" when they should not be. Examples include sentences embedded within a larger paragraph, bullets in a list-like enumeration, or text that introduces such lists.

\subparagraph{a2. Revision of Long Paragraphs.}
\label{par:artext-a2}
arText claro recommends avoiding overly long paragraphs—marked for revision by the system when they exceed 135 words \citep{da2022sistema}—and addressing a different topic in each paragraph. In the texts analyzed in this paper, there were few paragraphs of such length. However, the example below show the clear need for this recommendation:
\begin{quote}
    En todo caso, e independientemente de cuáles hubieran sido las rentas e ingresos del ejercicio anterior de la persona que vive sola o de la unidad de convivencia, cuando el solicitante del ingreso mínimo vital o uno o varios de los miembros de la unidad de convivencia, en su caso, tuvieran reconocida en la fecha de la solicitud, o les fuera reconocida antes de la resolución, una o más pensiones, contributivas o no contributivas, del sistema de la Seguridad Social, o un subsidio de desempleo para mayores de 52 años, cuyo importe mensual conjunto, incluida la parte proporcional de pagas extraordinarias, fuera inferior a la cuantía mensual de renta garantizada aplicable, y procediera el reconocimiento del ingreso mínimo vital por concurrir todos los requisitos para ello, el importe mensual de esta prestación no podrá ser superior a la diferencia entre la referida cuantía mensual de la renta garantizada y el importe mensual de la pensión o de la suma de las pensiones, incluida en su caso la parte proporcional de las pagas extraordinarias.
\end{quote}

As can be observed, the example paragraph also fails to comply with recommendation a1, as well as with others related to appropriate sentence length that will be discussed later in this section. To make this paragraph acceptably clear according to the system, several shorter sentences were needed, along with the deletion of repetitions and the use of more direct expressions. The result of this simplification is as follows:
\begin{quote}
    Es posible que al solicitante del IMV o miembro de una unidad de convivencia se le reconozca una asignación económica antes de resolverse la solicitud. Puede tratarse de un subsidio de desempleo para mayores de 52 años o una pensión, contributiva o no contributiva, de la Seguridad Social. Si esto ocurre, su importe mensual, incluida la parte proporcional de pagas extraordinarias, contará para el cálculo de los ingresos mensuales conjuntos. Esto se hará independientemente de las rentas e ingresos del ejercicio anterior. Por tanto, el importe mensual del IMV será la diferencia entre la cuantía mensual de renta garantizada aplicable y el importe mensual conjunto, incluidas las pensiones.
\end{quote}

In this recommendation, similar to the previous one, arText claro sometimes encounters issues, such as identifying separate paragraphs as a single one.
\subparagraph{a3. Introduction of connectors at the beginning of paragraphs.}
\label{par:artext-a3}
Here, arText claro recommends beginning paragraphs with discursive connectors as an explicit marker to link them effectively to the preceding paragraph. This recommendation does not seem to apply well to web texts. Since these are hypertexts and the reader does not necessarily access sequentially, the connection to paragraphs in previous sections becomes less relevant, as it cannot be assumed that the audience will read the entire content. Instead, each portion of text is under a subtitle that summarizes its connection to the overall topic.

Additionally, the system marks fragments that are not paragraphs under this recommendation. Furthermore, we observed that clicking on the title of the recommendation to expand the entire text sometimes results in the alterations of the introduced text, as some line breaks get automatically removed.

\subparagraph{a4. Revision of Long Sentences.}
\label{par:artext-a4}
This requirement is the most frequent and one of the most difficult to fulfil when following the recommendations provided by arText claro. A sentence is considered long if it exceeds 25 words \citep{da2022sistema}. However, in the administrative field, it is often necessary to mention institutions with very long names, causing a simple sentence to exceed this threshold. Examples include \textit{las Haciendas Tributarias Forales de Navarra y de los territorios históricos del País Vasco} (the tax offices of Navarre and the historical territories of the Basque Country) or \textit{el registro de mediadores sociales del ingreso mínimo vital} (the register of social mediators of the minimum vital income).

In some cases, we used lists to reduce sentence length when the typology allowed it. For example:
\begin{quote}
    Para las prestaciones por Incapacidad Temporal en pago directo del INSS, Nacimiento y cuidado de menor, Riesgo durante el embarazo, Riesgo durante la lactancia natural y Cuidado de menores afectados por cáncer u otra enfermedad grave, este certificado integral de prestaciones, sólo mostrará los datos relativos al certificado de IRPF.
\end{quote}

This sentence was simplified as follows:
\begin{quote}
    Este certificado integral de prestaciones sólo mostrará los datos relativos al certificado de IRPF para las siguientes prestaciones:
\begin{itemize}
    \item Incapacidad Temporal en pago directo del Instituto Nacional de la Seguridad Social (INSS)
    \item Nacimiento y cuidado de menor
    \item Riesgo durante el embarazo
    \item Riesgo durante la lactancia natural
    \item Cuidado de menores afectados por cáncer u otra enfermedad grave
\end{itemize}
\end{quote}

There are instances where the recommender cannot distinguish between two sections of text that do not form a single sentence. For example, when section headings do not end with a full stop, arText claro fails to identify them as separate from the following sentence, leading to incorrect marking of these sections as long sentences.

\subparagraph{a5. Division of Long Sentences.}
\label{par:artext-a5}

This recommendation refers to the same topic as the previous one, and presents the same difficulties. The difference is that a5 focuses mainly on compounds clauses. arText claro suggests dividing them in two or more sentences, depending on the composition. 

For example, here it is a sentence marked by arText claro for this recommendation:
\begin{quote}
    Este servicio permite solicitar la prestación por Nacimiento y Cuidado de Menor, para disfrutar del periodo de descanso laboral correspondiente, por nacimiento, adopción, guarda con fines de adopción y acogimiento permanente o temporal.
\end{quote}
It was divided in two:
\begin{quote}
    Este servicio permite solicitar la prestación por Nacimiento y Cuidado de Menor. Con ella se podrá disfrutar del periodo de descanso laboral correspondiente, por nacimiento, adopción, guarda con fines de adopción y acogimiento permanente o temporal.
\end{quote}

\subparagraph{a6. Variation of Connectors.}
\label{par:artext-a6}
The system suggests avoiding the repetition of connectors. This recommendation appears to be more a matter of style than a commitment to greater clarity. In fact, the analysis and classification of the main recommendations on clear language in Spanish within the legal-administrative field, conducted by \citet{daCunha_Escobar_2021}, do not include recommendations on this matter. An example of the suggestions made by arText claro is substituting \textit{si} (if) with phrases like \textit{en caso de} (in case), \textit{en el caso de que} (in the event that) or \textit{siempre que} (provided that).

\subparagraph{a7. Including Lists.}
\label{par:artext-a7}
arText claro suggests that well-constructed lists can be highly effective in conveying information clearly. However, the recommendation does not specify the criteria for a well-constructed list. 

Although our documents included some enumerations that could have been identified by the system, this recommendation was not offered for any of our texts. It is important to note, however, that lists were occasionally used to reduce sentence length, in line with recommendation a4.

\paragraph{b. Recommendations for the Morphosyntactic level}\textcolor{white}{.}
\label{par:artext-b}
\subparagraph{b1. Use of the passive voice.}
\label{par:artext-b1}
The system suggests using active voice as preferable for texts intended for citizens. We followed this recommendation by replacing periphrastic passive sentences with active sentences when the agent could be inferred from the context. Otherwise, we used reflexive passives, as illustrated in the following example. The fragment of the text highlighted in yellow corresponds to the markings provided by arText claro:

\begin{quote}
    La fe de vida deberá ser presentada en la Dirección Provincial del INSS que gestiona su pensión, información que \colorbox{yellow}{ha sido comunicada} a los pensionistas en la notificación de la revalorización de la pensión, durante el primer trimestre de cada año.
\end{quote}
The simplified version of this paragraph is as follows:
\begin{quote}
    La fe de vida deberá ser presentada en la Dirección Provincial del Instituto Nacional de la Seguridad Social (INSS) que gestiona su pensión. Esta información se ha comunicado a los pensionistas en la notificación de la revalorización de la pensión, durante el primer trimestre de cada año.
\end{quote}

It is noteworthy that in the presented example, arText claro identified only the second construction in passive voice, \textit{ha sido comunicada} ([this information] has been communicated), while the first one, \textit{deberá ser presentada} ([the proof of life] must be submitted), was not recognized as passive and, therefore, was not transformed into active voice in the simplified version. 

\subparagraph{b2. Revision of Gerunds.}
\label{par:artext-b2}
arText claro recommends avoiding the use of gerund verbs as they can lead to ambiguity. The example provided in their recommendation illustrates a clear instance where the attribution of responsibility for the action may be ambiguous. While the instances in our texts were not as noticeably ambiguous, we adhered to this recommendation and sought alternative linguistic forms. Here is an example:
\begin{quote}
    Se considerará pareja de hecho, a estos efectos, la constituida con análoga relación de afectividad a la conyugal con al menos dos años de antelación, por quienes, no \colorbox{yellow}{hallándose} impedidos para contraer matrimonio, no tengan vínculo matrimonial con otra persona y hayan convivido de forma estable y notoria con carácter inmediato a la solicitud de la prestación y con una duración ininterrumpida no inferior a cinco años.
\end{quote}
The proposed solution is as follows:
\begin{quote}
    Para considerarse pareja de hecho, a estos efectos, se requiere una relación de afectividad análoga a la conyugal con al menos dos años de antelación. Las personas que conformen la pareja no deben hallarse impedidas para contraer matrimonio, ni tener vínculo matrimonial con otra persona. Además, deben haber convivido de forma estable, notoria e ininterrumpida durante al menos los cinco años anteriores a la solicitud de la prestación.
\end{quote}

\subparagraph{b3. Revision of Participles.}
\label{par:artext-b3}
arText claro advises against the use of verbs in the participle form unless absolutely necessary, as they may "cause ambiguity, lengthen sentences, and make comprehension more difficult". Although the problem is somewhat vaguely stated, the example provided by the system offers some insight into the issue of identifying the agent of the action associated with participles. Nonetheless, it would be preferable if the term "subject" were not used\footnote{The syntactic function of the subject in the case of participles does not always refer to the agent performing the action (see: \href{https://www.rae.es/gramática/sintaxis/el-participio-características-fundamentales\#27.8j}{https://www.rae.es/gramática/sintaxis/el-participio-características-fundamentales\#27.8j} [Accessed: 28th June 2024]).}.
Here it is an example of the texts of our dataset:

\begin{quote}
    \colorbox{yellow}{Finalizado} el proceso de cumplimentación se le indicará los documentos que debe presentar y tendrá la posibilidad de adjuntarlos electrónicamente.
\end{quote}
This is the solution proposed:
\begin{quote}
    Cuando usted finalice el proceso de cumplimentación, se le indicarán los documentos que debe presentar y tendrá la posibilidad de adjuntarlos electrónicamente
\end{quote}

It can be observed in the given example that the ambiguity associated with the use of the participle is resolved by making the agent explicit as \textit{usted}, the formal variant of the 2nd person singular pronoun, rather than by altering the verb form. For instance, the use of the participle could have been replaced with an intransitive construction of the verb: \textit{Cuando el proceso de cumplimentación finalice, [...]}. In this case, the verb would have conveyed the meaning "to be completed", despite not displaying a passive structure, instead of "to complete something". This would not have resolved the ambiguity regarding who is responsible for completing the process.

Sometimes, the system marks words that are not participles but resemble them morphologically, such as \textit{requisito} (requirement), similar to \textit{escrito} (written). Additionally, it consistently marks all participles for review, even though many of them do not lead to a lack of clarity, as it was discussed in the previous recommendation. When it is suggested that all participles in the text should be revised, the user may struggle to identify which ones are in fact problematic. 
One of the motivations for developing a technological writing assistant was the difficulty that administrative staff experienced in applying the plain language recommendations from existing guides and manuals due to their unfamiliarity with linguistic terminology \citep{da2022sistema}. If this recommendation tends to be similarly challenging to apply, it will not fulfill its intended purpose effectively.

\subparagraph{b4. Elimination of Archaic Verb Forms.}\label{par:artext-b4}
arText claro recommends replacing outdated verb forms with simpler, more contemporary alternatives. The future subjunctive is a residual verb form in Spanish, mostly used in the legal-administrative field. In common language, the preterit imperfect subjunctive is used instead, making it the preferable verb form. Here is an example where we obtained this recommendation:
\begin{quote}
    La inexistencia de vínculos de parentesco entre todos o parte de los convivientes cuando uno de ellos \colorbox{yellow}{solicitare} el ingreso mínimo vital.
\end{quote}
And here is the simplification with an alternative verb form:

\begin{quote}
    La inexistencia de vínculos de parentesco entre todos o parte de los convivientes cuando uno de ellos solicitase el IMV.
\end{quote}

\subparagraph{b5. Systematic Use of First Person Verbs.}\label{par:artext-b5}
The system recommends the user to consistently refer to the sender in either singular or plural form throughout the text to maintain systematicity. It marks in different colours the units that it identifies as verbs in the 1st person singular or in the 1st person plural.

The use of the 1st person plural in out texts is systematic. However, arText claro marks as 1st person singular verb forms that actually are 3rd person. In Spanish, it is very common for the 1st and 3rd person singular to share the same verb form in the subjunctive or imperfect preterit indicative tenses. To illustrate this, the verb form \textit{pueda} (subjunctive form of the verb "can") could refer to \textit{yo pueda} (1st person singular) as well as \textit{él/ella pueda} (3rd person singular), or even \textit{usted pueda} (formal form of 2nd person singular).

\subparagraph{b6. Revision of Verbal Nominalization.}\label{pr:artext-b6}
The system recommends replacing nouns derived from verbs, unless they are terms specific to the field that cannot be altered, with their corresponding verbs to make the text easier to understand and more dynamic. According to the arText claro documentation \citep{da2022sistema}, the system identifies nouns starting with a lowercase letter that end in \textit{-ción} or \textit{-ciones}, except for terms commonly used in the administrative field included in an exclusion list. One example from the texts of our dataset is:
\begin{quote}
La residencia legal en España se acreditará mediante la \colorbox{yellow}{inscripción} en el registro central de extranjeros, en el caso de nacionales de los Estados miembros de la Unión Europea, Espacio Económico Europeo o la Confederación Suiza, o con tarjeta de familiar de ciudadano de la Unión o \colorbox{yellow}{autorización} de residencia, en cualquiera de sus modalidades, en el caso de extranjeros de otra nacionalidad.
\end{quote}
In the simplification, the word \textit{inscripción} (registration) is substituted, while \textit{autorización} (autorization) is maintained, because of being part of a fixed concept \textit{autorización de residencia} (residence authorization):
\begin{quote}
    La residencia legal en España se acreditará:
    \begin{itemize}
        \item En el caso de nacionales de los Estados miembros de la Unión Europea, Espacio Económico Europeo o la Confederación Suiza, al inscribirse en el registro central de extranjeros.
        \item En el caso de extranjeros de otra nacionalidad, mediante la tarjeta de familiar de ciudadano de la Unión o autorización de residencia, en cualquiera de sus modalidades
    \end{itemize}
      
\end{quote}

However, this recommendation presents a similar issue to the revision of participles (b2): often, the highlighted noun does not obstruct clarity or unnecessarily lengthen the sentence. This can make the recommendation challenging for users to interpret. Additionally, there may be no suitable verb replacement without altering the sentence's meaning, as in the case of \textit{situación} (situation) in contexts like this: \textit{[…] en atención a la situación personal de la víctima} ([...] in view of the victim's personal situation). Automatically identifying only problematic nominalizations remains a challenge.

\subparagraph{b7. Rephrasing ideas expressed in the negative.}\label{pr:artext-b7}
This arText claro recommendation suggests using affirmative formulations when possible, providing a clear example of how combining multiple negations in a sentence can obscure clarity. While negations in our texts are not complex and sometimes essential for explanations, we tried to adhere to this recommendation whenever possible. An example of this is:
\begin{quote}
    Los datos introducidos en el formulario deben coincidir con los existentes en nuestras bases de datos. Si \colorbox{yellow}{no} son coincidentes, \colorbox{yellow}{no} será posible obtener el certificado.
\end{quote}
The affected sentence is formulated in the positive:
\begin{quote}
    Los datos que introduzcas en el formulario deben coincidir con los existentes en nuestras bases de datos para que sea posible obtener el certificado.
\end{quote}
\paragraph{c. Recommendations for the Lexical Level}\textcolor{white}{.}
\label{par:artext-c}
\subparagraph{c1. Use of Subjectivity Indicators.}
\label{par:artext-c1}
This recommendation highlights units that, according to \citet{otaola1988modalidad}, may indicate subjectivity. artText claro advises users to review these units since administrative texts are typically expected to be objective. This recommendation was not suggested to our texts.

\subparagraph{c2. Introduction of acronyms.}\label{pr:artext-c2} 
The system marks text units structured as acronyms when they are not accompanied by their full forms, as it is recommended to include the full term at least the first time an acronym appears in the text. It specifically detects "proper acronyms" \citep{giraldo2008analisis}, which are formed solely from the initials of the lexical units they represent. In the analysed texts, the system failed to detect some acronyms that were not clarified by their full form at any point in the text.

We encountered instances where the full term of an acronym is less informative than the acronym itself, particularly for terms borrowed from English. Examples include \textit{SMS (Short Message Service)} and \textit{PDF (Portable Document Format)}. In such cases, it is preferable to replace the acronym with a Spanish synonym if possible, or provide a brief explanation of the concept. For instance, the Spanish version \textit{mensaje de texto} could be used instead of \textit{SMS}, and \textit{formato estándar de los documentos digitales} to explain \textit{PDF}.

Additionally, we identified two instances where the system incorrectly flagged units as acronyms. The first case is \textit{VIVESS}, the proper name of an administrative application. The second case is \textit{APP}, a shortening of \textit{application}, which misled the system because of being capitalized.

Finally, in our documents, some acronyms are accompanied by their corresponding full terms without being explicitly preceded by them in the text. Thanks to the digital format, this relationship is expressed through features such as displaying the acronym's meaning when hovering over it. For example, in HTML, this is represented as:

\begin{verbatim}
<acronym lang="es" title="Número de 
Identificación Fiscal">NIF</acronym>
\end{verbatim}

However, since arText claro only processes plain text, it detects the acronym but not its clarification, marking it for review. Although this strategy effectively takes advantage of the medium in which the text is presented, arText claro's recommendation remains advisable. Some readers may recognize the dashed line under the acronym as an indicator of additional information, which could lead them to miss the explanation.

\subparagraph{c3. Systematic use of acronyms.}
\label{par:artext-c3}
This recommendation advises that once an acronym is introduced in a text, the acronym should be used consistently instead of the full form. We observed an instance where the acronym \textit{CEA} appears along with its clarification \textit{Código Electrónico de Autenticidad} (Electronic Authenticity Code), but not at its first appearance in the text. The system should have flagged this as a lack of systematicity, but it failed to detect it.

Considering the nature of hypertexts, where readers can access the information in a non-sequential way, they may encounter the acronym before its full form. Therefore, although we adhered to this recommendation in our simplifications, we support the strategy of including the clarification in the HTML explained in the previous recommendation, as it can aid readers without making the text excessively long and repetitive, despite the drawbacks mentioned.

\subparagraph{c4. Use of More Transparent Terms.}
\label{par:artext-c4}
This recommendation suggests replacing words that, while not unfamiliar in common language, belong to a higher register that may be challenging for many readers. For instance, the verb \textit{efectuar} (to effect) is recommended to be replaced by its synonym \textit{realizar} (to make). However, some suggestions from the system do not always yield successful results; for example, it proposes replacing \textit{anverso} (obverse) with \textit{cara} (side), whereas we found it clearer to replace the entire expression \textit{anverso y reverso} (obverse and reverse) with \textit{ambas caras} (both sides). Additionally, certain words belong to larger units of meaning, such as \textit{débito} (debit) in \textit{tarjeta de débito} (debit card), where substituting it with \textit{deuda} (debt) would not be appropriate.

\subparagraph{c5. Substitution of Expressions That Are Difficult to Understand.}
\label{par:artext-c5}
arText claro suggests substituting expressions that hinder clarity because they are not commonly used. These expressions are identified from a predefined list and the system recommends clearer alternatives or adding brief explanations of their meaning. For example, it suggests using \textit{según} (according to) instead of \textit{de acuerdo con} (in accordance with) and \textit{para} (for) instead of \textit{a efectos de} (for the purpose of). However, some variations of these expressions are not detected by the system. An example is \textit{a cuyo efecto} (for which purpose), which may pose similar difficulties for readers, but no alternative is suggested.

\subparagraph{c6. Substitution of Inaccurate Words.}
\label{par:artext-c6}
arText claro recommends eliminating or substituting words that evoke a lack of accuracy. In our texts, the word \textit{hecho} (fact) is frequently marked because the system fails to recognize its part in the larger term \textit{pareja de hecho} (common-law partner). Similarly, the verb \textit{hacer} (make/do) is marked as vague by the system. However, in instances where other recommendations suggest splitting lengthy sentences, using this kind of verb can be helpful to avoid repeating the more precise verb, thereby maintaining conciseness and clarity.
\subparagraph{c7. Elimination of Redundant Expressions.}
\label{par:artext-c7}
The system identifies redundant expressions from a predefined list and advises removing them. However, this recommendation was not suggested in the simplification of the texts of our dataset.

\subparagraph{c8. Revision of Long Words.}
\label{par:artext-c8}
arText claro suggests shorter alternatives for some long words included in a predefined list \citep{sierra2022palabras} to enhance readability. For instance, the word \textit{gratuita} (free) is suggested to be replaced by its synonym \textit{gratis}. However, we noticed that the system sometimes marks a word for revision while ignoring the same word later in the text, which indicates that the detection is not systematic.

\section{LengClaro simplifications}
\label{sec:LengClaro-simplification}

To develop the second simplified version, we followed these steps:
\begin{enumerate}
\item We reviewed the recommendations compiled in the work of \citet{daCunha_Escobar_2021}\footnote{The sources analyzed were: 
\citet{vilches2010sarmiento, comision2011informe, comision2015europea, yanez2016, carretero2017lenguaje, aytomadrid}; and \citet{carretero2019comunicacion}.
}. While only the most frequent recommendations were included in arText claro, other recommendations from their research remain of interest.
\item We consulted additional guides and manuals on plain language in written Spanish within the legal-administrative field, published over the last five years (2019–2023) to identify any new contributions that warrant consideration\footnote{The sources consulted are: \citet{carretero2019fuentes, aragon2020estilo, comunidadmadrid, jurista2021GonzSoria, murcia2022}; and \citet{propuesta2023pobleteyepes}}. We also took into account recommendations specific to online texts\footnote{\citet{remediavagos}}. 
\item We analyzed the appropriateness of all these recommendations for texts published on web portals and selected the suitable ones.
\item Similar to the approach used in arText claro, we classified the recommendations into categories to build an organized set of rules.
\item Finally, we followed these rules to adapt the original documents for the second simplified version, referred to by the same name as our dataset: LengClaro.
\end{enumerate}

To avoid repeating the recommendations already discussed in Section \ref{sec:artext-claro}, we note that, with some exceptions, we generally adhered to the arText claro guidelines. In this section, we will briefly summarize them, while focusing on the differences in their implementation in the LengClaro versions, which are no longer constrained by the technical issues previously mentioned. Additionally, We will introduce new additions to the list of recommendations.

The recommendations are outlined below, organized by category.

\paragraph{a. Recommendations for the Discourse Level}\textcolor{white}{.}
\label{par:lengclaro-a}

We adhered to the following arText claro recommendations:
                    
\begin{itemize}
    \item Ensure paragraphs are not too long, address a single topic and contain at least two sentences (a1, a2).
    \item Use short sentences (a4, a5).
    \item Use well-constructed lists where appropriate (a7).
\end{itemize}

Operations concerning recommendation a1 were implemented but not as strictly as required by the writing assistant. Our texts are intended for the web, and thus for screen reading. In this specific context, "a sentence is an excellent paragraph", in \citet{remediavagos}'s words.

Concerning recommendations a4 and a5, we aimed to keep sentences under 25 words, while allowing for slightly longer sentences in certain instances. This is particularly relevant when dealing with sentences that include institutions with very long names, as outlined in recommendation a4. In addition, we observed that sentences containing a single subordinate clause easily exceed 25 words, despite being concise in content\footnote{We specifically refer to sentences containing only one subordinate clause, as the concatenation of multiple subordinate clauses has been widely identified as a source of ambiguity and lack of clarity. \citep{comision2011informe, yanez2016, carretero2017lenguaje, carretero2019comunicacion, carretero2019fuentes, jurista2021GonzSoria}}. Our perception is that splitting such sentences can sometimes lead to increased repetition and a less clear connection between elements that were originally linked in a single sentence.

According to \citet{irekia}, there are neurological differences when processing linear structures and hierarchical structures in a sentence. Hierarchical structures are processed more efficiently and, therefore, more effortlessly than linear structures. This indicates that, while sentence length does have an impact, it must be analysed together with the typology of the propositions it contains. According to \citet{dols2018, lentz2017, MaatDekker2016} and \citet{renkema2011}, as cited by \citet{van2024clearer}, sentence length is often used to assess how easily a text can be read, which refers to readability. However, readability and comprehensibility are not the same \citep{van2024clearer}. Comprehensibility refers to not only how easy a text is to read but also to how well a reader can grasp a text's meaning \citep{lentz2017, gavora2012text, renkema2011}.

Regarding recommendation a7, we established the following criteria for well-constructed lists in the LengClaro simplifications:

\begin{itemize}
    \item Lists should be preceded by an statement that introduces the elements that will comprise the list \citep{aytomadrid}
    \item Lists should be typographically highlighted on separate lines \citep{comision2011informe}
    \item The items in the list should be consistent in form, whether they are nouns, verbs, etc. \citep{comision2011informe, aytomadrid, aragon2020estilo, comunidadmadrid}. It has been observed that some lists in the original documents lacked this consistency, as demonstrated in the following example:
    \begin{quote}
    Para obtener un certificado:
    \begin{itemize}
        \item Una vez identificado, podrás obtener el certificado pulsando en el enlace "Certificado integral de prestaciones".
        \item Disponer del software para descargar/imprimir el certificado (archivo PDF).
        \item Si accede como representante, debe solicitar al representado que confirme la representación accediendo al enlace que recibirá en su móvil por SMS.
        \item Si accede como apoderado, deberá estar inscrito en el Registro Electrónico de Apoderamientos.
    \end{itemize}
\end{quote}
    \item List items's markers should be clear and uniform \citep{carretero2019fuentes, aytomadrid, aragon2020estilo, comunidadmadrid}. Although some sources recommend using numbers for enumerations and letters for alternating options, we followed the guidelines of \citet{remediavagos}, which are specifically oriented toward web writing. We used bullet-points to mark items in the lists, indicating enumeration or alternation in the introductory statement, and numbers only when the order was important, such as in a sequence of steps. Letters were used solely to mark items when they referred to different possible cases, where the list item itself constituted a paragraph.
\end{itemize}

For this category, some recommendations from arText claro were excluded. The first is the use of connectors at the beginning of paragraphs (a3). This recommendation was not implemented in our simplifications due to the hypertextual nature of the documents in our dataset. While connectors are essential for maintaining coherence, this exclusion does not imply that we avoided using them. It simply means that we focused on connecting ideas within the same topic, instead of connecting paragraphs, as the topics in our documents are organized into relatively independent sections. The second recommendation excluded is the use of varied connectors (a6), as it is primarily a stylistic consideration.

Within the discourse category, we list below additional recommendations that we also followed:

\subparagraph{a8. Avoid lengthy "copy and paste" fragments.}
\footnote{We maintain the nomenclature used in Section \ref{sec:artext-claro} to identify the recommendations and continue the numbering from where it left off.}
\label{par:lengclaro-a8}
\citep{comision2015europea, montolio2019hacer, murcia2022}.

It is not advisable to recycle an existing text if it is not well adapted. When fragments from other documents are to be incorporated into a text, it is crucial to adjust them to their new linguistic context. Ensuring consistency and avoiding repetitions, omissions, or contradictions is essential, as these issues can negatively impact the internal logic and clarity of the final text \citep{comision2015europea}. For example, in the Document 1 on the MVI, we observed that the figures for the guaranteed income amounts had been updated, but those for the financial requirements were still the same as the previous year, despite they are correlated.

\subparagraph{a9. Get to the point.}
\label{par:lengclaro-a9}
\citep{remediavagos, murcia2022, propuesta2023pobleteyepes}

This recommendation is particularly motivated by the the fact that our documents are designed for a portal web. On the web, users do not read in the traditional way; instead, they scan the text on the screen \citep{aragon2020estilo}, searching for clues about the content of each section before deciding to read it. As a result, slow introductions are less effective in this context \citep{remediavagos}.

To address this, we have placed the most relevant and useful information at the beginning of each paragraph \citep{remediavagos}, allowing readers to quickly identify the main idea \citep{murcia2022, propuesta2023pobleteyepes}. Research has shown that readers tend to understand and retain information better when paragraphs begin with the main topic \citep{remediavagos}. For example, in the text below, the main idea (in bold) is centered on the non-payment of the debt and its consequences:
\begin{quote}
    En los supuestos anteriores, transcurrido el plazo de ingreso en periodo voluntario \textbf{sin pago de la deuda}, se aplicarán los correspondientes recargos y comenzará el devengo de intereses de demora, sin perjuicio de que estos últimos solo sean exigibles respecto del período de recaudación ejecutiva. En los supuestos que se determinen reglamentariamente, la entidad gestora podrá acordar compensar la deuda con las mensualidades del ingreso mínimo vital hasta un determinado porcentaje máximo de cada mensualidad.
\end{quote}

It was simplified as follows:
\begin{quote}
    \textbf{Si no se paga la deuda} dentro del plazo en periodo voluntario, se aplicarán los recargos e intereses de demora correspondientes al periodo ejecutivo. En ciertos casos que se determinen reglamentariamente, la entidad gestora podrá compensar la deuda con las mensualidades del IMV hasta un determinado porcentaje máximo de cada una.
\end{quote}

\paragraph{b. Recommendations for the Morphosyntactic level}\textcolor{white}{.}
\label{par:lengclaro-b}

We adhered to all of arText claro's morphosyntactic recommendations, albeit with some nuances. In summary, these include:

\begin{itemize}
    \item Express the agent of the action whenever possible, avoiding passive constructions and non-personal verb forms such as gerunds and participles (b1, b2, b3).
    \item Eliminate archaic verb forms (b4).
    \item Maintain consistency in the use of the 1st person (b5).
    \item Replace verbal nominalizations with their corresponding verbs when possible (b6).
    \item Avoid unnecessary negative formulations (b7).
\end{itemize}

Regarding recommendation b1, in the lengClaro version we did not have to face the system's issue with recognizing passive voice in more complex structures, such verbal periphrasis with a passive construction as the main verb. Where feasible, we also replaced standard reflexive passive forms with active constructions. Moreover, it is interesting to mention that there is a passive construction whose excessive use is characteristic of legal-administrative language: the reflexive passive with an explicit agent \citep{de2000texto}. In this construction, the passive with the morpheme \textit{se} appears accompanied by the agent preceded by the prepositions \textit{de} (of) or \textit{por} (by). This syntactic structure is not permitted in common language, therefore it is not advisable to use it when the text's audience is not familiar with it. Consequently, we replaced this type of passive construction, which was already discouraged by \citet{comision2011informe}, as noted by \citet{daCunha_Escobar_2021}. Below is an example:
\begin{quote}
    Los requisitos de ingresos y patrimonio para el acceso y mantenimiento de la prestación económica de IMV \textbf{se realizará por la entidad gestora} conforme a la información que se obtenga por medios telemáticos de la Agencia Estatal de Administración Tributaria y en las Haciendas Tributarias Forales de Navarra y de los territorios históricos del País Vasco
\end{quote}
The reflexive passive with an explicit agent was replaced by an active sentence:
\begin{quote}
    Los requisitos de ingresos y patrimonio. Para acreditarlos, \textbf{la entidad gestora recurrirá} a la Agencia Estatal de Administración Tributaria y a las Haciendas Tributarias Forales de Navarra y de los territorios históricos del País Vasco.
\end{quote}

With regard to the recommendations on non-personal verb forms (b2, b3), we refocused our approach on the root of the ambiguity: the inability to assign responsibility for an action. Rather than solely avoiding these specific verb forms, we targeted how they create ambiguity in the text. As illustrated in recommendation b3, merely using a conjugated verb may not always resolve the issue. Therefore, instead of avoiding non-personal verb forms entirely, we aimed to make the agent of the actions explicit. We avoided gerunds except when they functioned as the main verb in a verbal periphrasis. For participles, we specifically concentrated on avoiding absolute constructions. Some sources, such as \citet{yanez2016}, as noted by \citet{daCunha_Escobar_2021}, or \citet{aragon2020estilo}, also advise against using infinitives as the main verb of a sentence, and we avoided that use too.

In addition to these recommendations, we adhered to the following suggestions in this category:

\subparagraph{b8. Address Readers Directly.}
\label{par:lengclaro-b8}
\citep{vilches2010sarmiento, comision2015europea, aragon2020estilo, propuesta2023pobleteyepes}

In some documents within our dataset, we observed an alternation between impersonal forms with \textit{se} and direct appeals to the reader when detailing instructions or steps to follow. Mentioned sources recommend the latter approach to engage the reader more effectively.

In addition, just as it is recommended to systematically use the first person, it is equally important to be consistent when addressing the reader. We also noticed instances of alternation between verbs conjugated in the 2nd person singular \textit{tú} and those of \textit{usted} (its formal variant) in some of our documents, which can lead to ambiguity.

Both forms are correct and can set either a closer or more formal tone, but it is crucial to use one consistently throughout the document. Based on the experience gained during the simplification process, we believe it is more practical to address the reader using the \textit{tú} form. This choice is preferable because its verb conjugations are unique to this grammatical person. In Spanish, the absence of an explicit subject is common, and since the verb forms corresponding to \textit{usted} are shared with the 3rd person singular, using the 2nd person singular allows us to avoid frequently specifying \textit{usted} to distinguish it from a potential 3rd person singular subject. In addition, the \textit{tú} form is widely used in the Spanish from Spain, without conveying disrespect, and it makes the text sound more approachable \citep{comunidadmadrid}. 

\subparagraph{b9. Avoid parenthetical remarks.} \citep{jurista2021GonzSoria, propuesta2023pobleteyepes}
\label{par:lengclaro-b9}

Parenthetical inserts disrupt the flow of reading. Instead of embedding additional comments within a sentence, it is preferable to develop these ideas in separate sentences, or, if they pertain to the entire sentence, to place them at the end \citep{carretero2019fuentes}. For example, the following sentence contains two parenthetical remarks:
\begin{quote}
    Mejora del porcentaje de la pensión de orfandad absoluta, que pasa del 52\% anterior al 70\% en los casos de carencia de rentas \textbf{(rendimientos inferiores al 75\% del salario mínimo interprofesional)} de los miembros de la unidad familiar de convivencia; e incremento del conjunto de pensiones de orfandad en el caso de existencia de varios beneficiarios, al permitirse alcanzar el 118\% de la base reguladora \textbf{(hasta entonces era el 100\%)} y establecerse una garantía de importe mínimo conjunto.
\end{quote}

To avoid such interruptions, the first remark is placed in a separate sentence, while the second is moved to the end of the sentence it applies to:
\begin{quote}
    Mejora el porcentaje de la pensión de orfandad absoluta:
        \begin{itemize}
            \item En los casos de carencia de rentas, este porcentaje pasa del 52\% al 70\%. Se considera carencia de rentas cuando \textbf{los rendimientos} de las personas integrantes de la unidad de convivencia sean \textbf{inferiores al 75\% del salario mínimo interprofesional}.
            \item Cuando haya más de una persona beneficiaria, se incrementa el conjunto de pensiones de orfandad. Dado este caso, se permite alcanzar el 118\% de la base reguladora, \textbf{que antes era el 100\%}. Además, se establece una garantía de importe mínimo conjunto.
        \end{itemize}
\end{quote}

\subparagraph{b10. Choose verb forms with precision.} \citep{aragon2020estilo, carretero2019fuentes}
\label{par:lengclaro-b10}
 
The verb form must align with the information being conveyed \citep{aragon2020estilo}. 
This should not need to be explicitly stated as a plain language recommendation; rather, it should be understood as a fundamental requirement in any form of writing. However, administrative texts frequently contain incorrect choices of verb forms. \citep{carretero2019fuentes}. In the documents of our dataset, we have primarily observed these errors in conditional constructions. For example: %tachado
\begin{quote}
    Los descendientes citados \textbf{podrán} ser hasta el segundo grado si no \textbf{\sout{estuvieran} están} empadronados con sus ascendientes del primer grado.
\end{quote}
\begin{quote}
    Estas medidas \textbf{pueden} adoptarse cautelarmente, [...] si la víctima \textbf{\sout{fuera} es} el sujeto causante de la prestación.
\end{quote}

\subparagraph{b11. Use the simplest syntax: subject + predicate.} \citep{vilches2010sarmiento, yanez2016, carretero2017lenguaje, aytomadrid, carretero2019comunicacion, carretero2019fuentes, aragon2020estilo, comunidadmadrid, jurista2021GonzSoria, propuesta2023pobleteyepes} 
\label{par:lengclaro-b11}

There is broad consensus in the literature supporting this recommendation. Spanish is an SVO (subject-verb-object) language, and this unmarked order is the most logical and easiest to interpret. This structure ensures that sentences flow more naturally and align with readers' expectations. Below is an example:

\begin{quote}
    Asimismo, estarán exentos del pago de tasas de expedición y renovación de Documento Nacional de Identidad lo menores de 14 años integrados en una unidad de convivencia que solicite la prestación de ingreso mínimo vital.
\end{quote}

In the simplified version, the subject was placed before the main verb:
\begin{quote}
    Los menores de 14 años integrados en una unidad de convivencia beneficiaria del IMV tampoco tendrán que pagar las tasas de expedir y renovar el DNI. 
\end{quote}
\paragraph{c. Recommendations for the Lexical Level}\textcolor{white}{.}
\label{par:lengclaro-c}

We adhered to the following arText claro's recommendations:

\begin{itemize}
    \item Avoid subjectivity (c1).
    \item Introduce acronyms the first time they appear in the text, and use the acronym consistently thereafter (c2, c3).
    \item Use terms and expressions that are as plain and easy to understand as possible (c4, c5, c8).
    \item Avoid redundant expressions (c7).
\end{itemize}

Regarding recommendation c1, we were able to avoid subjectivity even when it was not explicitly conveyed through traditionally subjective units. For example, we chose not to use the adjective sencillo (simple) in the following instance:

\begin{quote}
    El funcionamiento de la APP es muy sencillo, y en tres pasos se podrá cumplir con el trámite de acreditación de la vivencia sin desplazamientos:
\end{quote}
Whether the app is simple to use is a matter of opinion, so we retained only the essential information:
\begin{quote}
    Para utilizar la aplicación móvil debes seguir los siguientes pasos:
\end{quote}

Recommendations c4, c5, and c8 in arText claro rely on closed lists for detecting simplifiable units. In the LengClaro versions, we addressed a broader range of difficult-to-understand terms and expressions. For instance, we simplified \textit{consignar datos} (record data) to \textit{proporcionar datos} (provide data) or \textit{proceso de cumplimentación} (process of completion) to \textit{rellenar} (to fill in).

The recommendation for the lexical level in aText claro that we excluded in LengClaro is the sustitution of inaccurate words (c6). The writing assistant marks for revision words such as hyperonymous verbs, indefinite determiners or indefinite pronouns. However, in some cases, during this second simplification process, using these generic words was a strategic choice to avoid repeating concepts mentioned earlier or later in the text, as recommended by \citet{comision2015europea} and  \citet{yanez2016}. Since accuracy was a priority in LengClaro simplifications, we aimed to maximize the precision of the terms used. For example, \textit{justificación} (justification) was replaced by \textit{justificante} (bank receipt) when the text referred to a document attesting to a payment\footnote{Despite their similarity in Spanish, these terms are used in different contexts according to \citet{diccMoliner}'s definitions.}.

Below, we present the additional recommendations we followed for the lexical category:

\subparagraph{c9. Eliminate superfluous words.} \citep{vilches2010sarmiento, comision2015europea, yanez2016, aytomadrid, aragon2020estilo, comunidadmadrid}
\label{par:lengclaro-c9}

\citet{aragon2020estilo} identify various types of words that add nothing but noise to the text, including metaconcepts, hackneyed phrases, and unnecessary pronouns. The latter frequently appear in our original texts, as in constructions like \textit{de las mismas} (of them) instead of simply using the possessive determiner. The phrase \textit{en su caso} (if applicable) also appears frequently in our original texts, as if it were not evident to the readers that if the mentioned circumstances do not apply to them, neither do their implications. In addition, \citet{comunidadmadrid} recommend eliminating supporting verbs. For example, in our original documents:
%\textit{podrá proceder [...] a la rectificación de errores} was replaced by \textit{podrá [...] rectificar errores}.
	
\begin{quote}
    El \textbf{acceso al servicio podrá hacerse} con cualquier sistema de identificación válido. Con este acceso accederá a la simulación de la prestación teniendo en cuenta la información que incorpore.
\end{quote}
The supporting verb and its accompanying noun were replaced with the corresponding direct verb:
 \begin{quote}
     Para realizar la simulación necesitas:
 \begin{itemize}
     \item \textbf{Acceder al servicio} con un sistema de identificación válido.
     \item Proporcionar la información y documentos que se requieran.
 \end{itemize}
 \end{quote}
 
\subparagraph{c10. Avoid foreign words.} 
\label{par:lengclaro-c10} 
\citep{yanez2016, carretero2019comunicacion, carretero2019fuentes, propuesta2023pobleteyepes}

Not all readers may be familiar with foreign terms. Their use may hinder comprehension and lead to misunderstandings or incorrect interpretations. It is advisable to use precise terms in the text's original language. If a foreign word must be used, it should be accompanied by a translation or explanation. Good examples are the issues presented in the analysis of recommendation c2: it is preferable to explain the concept of an acronym rather than presenting its full form in English. 

\subparagraph{c11. Use inclusive language.} 
\citep{aytomadrid}
\label{par:lengclaro-c11} 

While this recommendation is not directly related to plain language principles, it plays a crucial role in the way in which the administration's communications reach citizens. As defined by the virtual platform for inclusive communication MODII in its guide to non-sexist language\footnote{\href{https://modii.org/wp-content/uploads/2022/01/lenguajenosexista.pdf}{https://modii.org/wp-content/uploads/2022/01/lenguajenosexista.pdf} [Accessed: 5th September 2024]}, this consists of :
\begin{quotation}
	%Comunicarnos de una manera que no represente una distinción asimétrica, desigual, excluyente o injusta entre mujeres, hombres y personas de genéro no binario.
    Communicating in a way that does not represent an asymmetrical, unequal, exclusionary, or unfair distinction between women, men and people of non-binary gender.
\end{quotation}

A common question that becomes particularly relevant for this work is whether it is possible to communicate in a manner that is both inclusive and clear\footnote{\href{https://modii.org/wp-content/uploads/2023/11/Guia-de-comunicacion-inclusiva-de-MODII-sobre-lenguaje-claro_accesible.pdf}{https://modii.org/wp-content/uploads/2023/11/Guia-de-comunicacion-inclusiva-de-MODII-sobre-lenguaje-claro\_accesible.pdf} [Accessed: 5th September 2024]}. This question arises when considering strategies that may conflict with certain recommendations for plain language. One such example is gender splitting, which often results in lengthier sentences, thereby making the text more cumbersome to read \footnote{\href{https://comunicacionclara.com/docs/Prodigioso-Volcan-habla-guia-comunicacion-inclusiva-2023.pdf}{https://comunicacionclara.com/docs/Prodigioso-Volcan-habla-guia-comunicacion-inclusiva-2023.pdf} [Accessed: 5th September 2024]}. A similar issue arises with explanatory appositions. Another example is the use of gender doublets using slashes or parentheses, which introduces accessibility problems by cluttering the text with non-normative elements. In some contexts, the use of symbols such as "@", "x", or "e" has become popular, but it results in a difficult or impossible pronunciation and, in digital communication, can cause confusion, particularly when "@" is interpreted as a symbol for email addresses or social media mentions. Choosing to use the generic feminine can lead to ambiguity. Other strategies, such as omitting explicit subjects or using impersonal forms compete with the above-mentioned recommendation to make the agent of the actions explicit.

Nevertheless, there are numerous strategies available for avoiding the use of the generic masculine without compromising clarity. Some of those suggested by \citet{aytomadrid} include the following:

\begin{itemize}
	\item Using generic or collective nouns.
	\item Using metonymic constructions.
	\item Using the imperative form. Other constructions that address the reader directly also avoid the generic masculine (\textit{Los pensionistas de la Seguridad Social[...].} $\rightarrow$ \textit{Si eres pensionista de la Seguridad Social[...].}).
	\item Omitting determiners where appropriate, or choosing determiners without gender marking such as \textit{cada}.
	\item Opting for relative pronouns such as \textit{quien} or \textit{quienes} instead of \textit{el que} or \textit{los que} respectively.
	\item Using periphrastic constructions (\textit{los beneficiarios} $\rightarrow$ \textit{las personas beneficiarias}).
\end{itemize}

Regrettably, it is not always possible to implement a strategy that simultaneously achieves both clarity and inclusivity. In such cases, clarity must take precedence, as applied in our simplified versions. We can observe this in the following example, in with the generic masculine forms are in bold:
\begin{quote}
    La residencia legal en España se acreditará [...] con tarjeta de familiar de \textbf{ciudadano} de la Unión o autorización de residencia, en cualquiera de sus modalidades, en el caso de \textbf{extranjeros} de otra nacionalidad. 
\end{quote}

The term \textit{familiar de ciudadano de la Unión} (family member of a European Union citizen) refers to a specific legal category and denotes a particular type of residence permit. In this case, for the sake of clarity, we opted to retain the masculine form. On the contraty, we replaced the term \textit{extranjeros} (foreigners):

\begin{quote}
    En el caso de personas de otra nacionalidad, se acredita mediante la tarjeta de familiar de \textbf{ciudadano} de la Unión o autorización de residencia, en cualquiera de sus modalidades.
\end{quote}

\paragraph{d. Recommendations for the Planning Phase}\textcolor{white}{.}
\label{par:lengclaro-d}

During the analysis and classification of the recommendations, it became apparent that the categories of discourse, morphosyntactic, and lexical are insufficient, as they only pertain to the operation of textualization. Textualization consists of transforming content into written language and represents one of the three phases that comprise the process of textual composition, as outlined by \citet{camps1990modelos}. The other two phases are planning and revision. The planning phase involves forming a mental representation of the information to be included in the text, while the revision phase involves reading and evaluating the text \citep{cassany1989describir}.

These phases are not linear but recursive \citep{camps1990modelos}. This is evidenced by the fact that the need of simplification results from the revision phase—in which it was determined that the text does not achieve the desired level of simplicity— in the re-composition process of a text. Achieving this simplification requires not only re-engaging with the textualization operation but also revisiting the planning phase, to amend the plan and rewrite the ideas accordingly.

Therefore, we also adhered to recommendations for the planning phase, including some that were already present in the manuals revised by \citet{daCunha_Escobar_2021} but were not included in their compilation. This phase involves drafting the text and serves three purposes: formulating objectives, generating ideas, and organizing them \citep{cassany1989describir}. In the formulation of objectives, the general features of the text are determined: the intention (textual typology), the target audience, the medium where the text will be read, etc.\footnote{\url{https://cvc.cervantes.es/ensenanza/biblioteca_ele/diccio_ele/diccionario/planificacionescrito.htm} [Accessed: 1st August 2024]} This contextualization is crucial for deciding what information the text should contain. In our case, these variables are predefined: expository texts for the general public on the Spanish Social Security portal web. Nonetheless, as recommended by \citet{carretero2019fuentes}, \citet{aragon2020estilo}, and \citet{comunidadmadrid}, we must keep them in mind to ensure the text appropriately fits its context.

Idea generation involves making an outline of the points to be developed. Since our work is a simplification of existing texts rather than starting from scratch, the ideas have already been generated during the original texts' composition.
However, these ideas can still be reorganized for a more coherent structure. We focused the recommendations of this category in this reorganization, as a well-structured message is crucial for clear language \citep{aragon2020estilo}. Below, we present the recommendations.

\subparagraph{d1. Group related ideas under headings.}
\label{par:lengclaro-d1} 
\citep{comision2015europea, remediavagos, aragon2020estilo, comunidadmadrid}

Organizing the ideas of a text involves establishing the parts it contains \citep{comunidadmadrid}. For the text to maintain coherence, it is essential to associate related ideas and assign each grouping a title. This recommendation is based on two primary reasons: first, information that is grouped into categories, ordered, and labeled is easier to understand \citep{aragon2020estilo}; second, headings facilitate the reading experience by enabling readers to quickly locate information and anticipate the content of each section \citep{comunidadmadrid}. To ensure that the structure remains clear and manageable, it is advisable to use no more than three levels of subheadings\citep{remediavagos}.

Nearly every document in our dataset was structured with sections and headings; however, the distribution of ideas was not always well-considered. Occasionally, the boundaries between sections were blurred. For example, original Document 7 contains two very similar and brief sections that might be better consolidated into a single section. These sections are: \textit{Quien puede solicitar la TSE?}, which literally means \textit{Who can request the EHIC?}, but the title for the English version is \textit{The EHIC can be requested}; and \textit{Solicitud y Renovación}, translated as \textit{Application and Renewal}.

\subparagraph{d2. Summarise the content.}
\label{par:lengclaro-d2} 
\citep{comision2015europea, aragon2020estilo, comunidadmadrid, jurista2021GonzSoria}

Starting a text with a brief summary of its content helps set the reader's expectations for what can be found within. Including a table of contents further enhances this by providing a detailed outline of the text's structure and guiding the reader in locating specific information. By listing all sections at the beginning of the document, readers can quickly identify the content covered without needing to scroll through the entire text. Additionally, as our texts are designed to be integrated into a web page, the table of contents can be navigable, allowing readers to jump directly to the sections of their interest.

\subparagraph{d3. Avoid repetition.}
\label{par:lengclaro-d3} 
\citep{comision2015europea, jurista2021GonzSoria, murcia2022}.

When readers encounter a repetition in a text without any indication that it serves as a reminder or emphasis, they often assume that it introduces new information. This assumption arises from the belief that if the author has taken the effort to write again a point, there must be some subtle difference between the instances. Consequently, readers may feel compelled to reread the passage to ensure that they have not missed any significant details. To illustrate how such repetitions can disrupt the clarity of a text, we present two pertinent examples. The first one involves a section that specifies the intended audience for the service described in Document 6. The two fragments in bold seemingly refer to the same set of benefits:

\begin{quote}
    A quién va dirigido 
    
    A personas que perciban \textbf{una prestación del Sistema de Seguridad Social o de una entidad ajena al Sistema pero integrada en el Registro de Prestaciones Sociales Públicas}, así como a personas que no perciban \textbf{una prestación de la Seguridad Social o de una entidad ajena al Sistema}.
\end{quote}

After deciphering the original paragraph, which states one condition and then its opposite, the reader concludes that it could have been summarized in one word: everybody. This is because the most important point—that the service is intended for anyone who wants to certify their status regarding the mentioned benefits—was omitted. In the LengClaro simplified version, this omission was corrected, and the repetition was replaced with a single mention:

\begin{quote}
    A quién va dirigido  
    
    A cualquier persona que necesite certificar que percibe o no percibe una prestación del Sistema de Seguridad Social. También se recogen las prestaciones de entidades ajenas al Sistema, pero integradas en el Registro de Prestaciones Sociales Públicas.
\end{quote}

The second example is even more blatant, as an entire new paragraph was added to address a particular condition that has no different consequence from the one described in the preceding paragraph. The repeated information is in bold:

\begin{quote}
\begin{itemize}
    \item Efecto de pago: \textbf{El pago efectuado mediante tarjeta de débito o de crédito se entenderá realizado en la fecha en que los fondos tengan entrada en la cuenta restringida de la Tesorería General de la Seguridad Social consignada al efectuar el ingreso}. Como regla general, entre las 24 y las 48 horas siguientes a la ejecución correcta de la operación en el terminal de conexión.

	Si por alguna circunstancia el ingreso correspondiente al pago efectuado con tarjeta de débito o de crédito se produjese transcurridas 48 horas, \textbf{el pago se entenderá  realizado en la fecha en que tal ingreso tenga entrada en la cuenta restringida de la Tesorería General de la Seguridad Social consignada en el momento de la realización del pago}.
 \end{itemize}
\end{quote}

We omitted the repetition without any loss of information, saving the reader time and effort in determining which situation applies to them. Although we required several readings to reach that conclusion with a sufficient degree of confidence, we inferred that the emphasis on the case of income arriving after 48 hours was only because it deviates from the general rule mentioned. Therefore, in the simplified version, we included a caveat to ensure this possibility does not go unnoticed:

\begin{quote}
    Cuándo una persona queda liberada de la deuda.
    
    El pago solo se considerará realizado cuando el dinero entre en la cuenta de la TGSS. Es en ese momento cuando la persona responsable del pago quedará liberada de la deuda. Por regla general, esto suele tardar entre 24 y 48 horas desde el momento en el que se realice el pago, aunque en ocasiones puede ser algún día más. Por ese motivo, es importante asegurarse de realizar el pago con tiempo de que los fondos lleguen a la TGSS dentro de plazo.

\end{quote}

\paragraph{e. Recommendations for the Revision Phase}\textcolor{white}{.}
\label{par:lengclaro-e}

\citet{aragon2020estilo} emphasize that revising the resulting text is essential to ensure it meets the objectives of simplicity. They recommend three strategies, that we followed during this process of simplification:

\begin{enumerate}
    \item Read the text aloud: Listening to the content offers a new perspective.
    \item Let the text rest and review it later: This helps in identifying errors or shortcomings that might have been overlooked.
    \item Have another person read the text: An external reader can provide fresh insights and feedback.
\end{enumerate}

Although more abstract, these recommendations should not be overlooked. In many cases, elements of ambiguity could be detected through a thorough and relaxed review of the documents.

\paragraph{f. Recommendations for Orthography}\textcolor{white}{.}
\label{par:lengclaro-f}

In addition to following the rules for correct spelling, which is essential for producing clear text, the manuals we reviewed offer additional recommendations that further enhance clarity at this level.

\subparagraph{f1. Limit the use of capital letters.}
\label{par:lengclaro-f1} 
\citep{carretero2019fuentes, aragon2020estilo,comunidadmadrid, murcia2022, propuesta2023pobleteyepes}

Excessive use of capital letters in a text can hinder readability. They should be used only when required by the rules, such as in proper names or acronyms. Special attention should be given to headings, as this is where capitalization is most often overused.

\subparagraph{f2. Use numerals instead of spelling out simple figures.}\textcolor{white}{.}
\label{par:lengclaro-f2}

Numbers with up to four digits are quickly identifiable and easier to read on screen. For larger numbers with several zeros, which can be confusing and difficult to interpret, it is preferable to indicate the number followed by the words "thousand", "million" or "billion". \citep{remediavagos}

\section{Discussion}
\label{sec:chap5-disc}
In this chapter, we detail the implications and insights derived from our research on the simplification of administrative texts into plain language. First, we expose potential improvements for the writing assistant arText claro based on our findings and experiences as users of the tool. Next, we address the practical and theoretical linguistic challenges that arise when simplifying complex administrative language, highlighting the obstacles encountered and the solutions applied. Finally, we evaluate preliminary experiments involving recent LLMs in the context of administrative text simplification, discussing their effectiveness and future potential in this field. 
%tus observaciones, lo que has aprendido (lo de los gerendios, el lenguaje inclusivo...)
\subsection{Proposals for improvement of arText claro}
\label{sec:artext-proposal}

During the simplification process it was observed that arText claro sometimes struggles with identifying paragraph and sentence boundaries, leading to incorrect marking of text fragments for revision. This issue was already mentioned by \citet{da2022redactor}. After our analysis, we identified some areas for improvement in the recommendations, which could benefit arText claro's users. The following adjustments are suggested:
\begin{itemize}
   \item \textbf{Recommendation a5}: Reconsider the word limit per sentence. Adhering to the 25-word limit stipulated by this recommendation can be particularly challenging in administrative texts. Especially in sentences containing one subordinate clause, our experience indicates that users often struggle to meet this limit, and after multiple attempts, the resulting sentences may display less cohesive logical relationships between the elements of the original sentence, without necessarily enhancing clarity. To make this recommendation easier for users to follow, rather than enforcing a strict 25-word maximum for each individual sentence, the average number of words per sentence across the entire document could be calculated. A desirable maximum average could be set at 20 words per sentence, for instance. This approach could be combined with a looser maximum word limit per sentence, allowing the system to continue flagging sentences that exceed an acceptable threshold, such as 35 words. This would ensure that no sentence longer than 35 words would be overlooked, while also permitting sentences longer than 25 words to be balanced by shorter ones, as long as the average sentence length remains within the recommended range. By relaxing this requirement, users could better focus on effectively reducing the length of the most problematic sentences while still adhering to the principles of the analyzed recommendations. The specific figures mentioned here are indicative; further empirical study would be necessary to establish these limits with greater precision and to ensure that elements like headings or list items do not artificially lower the average sentence length, potentially misrepresenting the overall clarity of the text.
   \item \textbf{Recommendation b1}: Expand the detection of passive constructions. Currently, arText claro is unable to identify more complex passive structures, such as reflexive passives with an explicit agent or passives where the verb \textit{to be} serves as the main verb in a verbal periphrasis. Enhancing the system's capabilities to detect these structures would represent a significant advancement, enabling it to suggest revisions to the user more effectively.
    \item \textbf{Recommendation b3}: Restrict the participles marked for revision. As mentioned previously, it would be clearer for the user if only participles that may induce ambiguity were marked for revision, rather than all participles. The examples in the recommendation suggest that the participles to be avoided are primarily those positioned at the beginning of a sentence, particularly when the agent of the action is unspecified. A possible solution is to exclude from the recommendation participles that are part of a conjugated verb, such as those following the verb \textit{haber} (to have). Additionally, participles functioning as adjectives, especially when they follow the noun that serves as the core of the nominal group and agree with it in gender and number, could also be excluded from user revisions.
    \item \textbf{Recommendation b6}: Focus on nominalizations accompanied by a complement introduced by \textit{de} (of). As in the previous case, it would be easier for the user to improve the text if not all the nominalizations were marked for revision. Based on observations during the text simplification process, we propose to focus recommendations on nouns most likely to be problematic. Nouns ending in \textit{-ción} typically derive from verbs and often express the action or effect of the corresponding verb\footnote{\href{https://www.rae.es/gramática/morfología/el-sufijo-ción-y-sus-variantes-ii-aspectos-semánticos-y-dialectales}{https://www.rae.es/gramática/morfología/el-sufijo-ción-y-sus-variantes-ii-aspectos-semánticos-y-dialectales} [Accessed: 10th June 2024]}. We observed that nouns with an action interpretation are more successfully replaced by verbs that nouns with an effect interpretation. However, both interpretations can occur, and distinguishing between them can be challenging \footnote{\href{https://www.rae.es/gramática/morfología/introducción-aspectos-generales-de-la-derivación-nominal\#5.1l}{https://www.rae.es/gramática/morfología/introducción-aspectos-generales-de-la-derivación-nominal\#5.1l} [Accessed: 10th June 2024]}. Therefore, to address this, we will focus on structural rather than semantic aspects. When a noun is accompanied by a complement introduced by the preposition \textit{de} (of), replacing it is simpler and beneficial for clarity. This is because verbal nominalizations can syntactically manifest, with certain restrictions, arguments that correspond to the verbs that constitute their lexical bases\footnote{\href{https://www.rae.es/gramática/sintaxis/las-nominalizaciones-i-sus-clases-nominalizaciones-de-acción-y-efecto}{https://www.rae.es/gramática/sintaxis/las-nominalizaciones-i-sus-clases-nominalizaciones-de-acción-y-efecto} [Accessed: 10th June 2024]}. For example, in the sentence \textit{La utilización de este servicio es totalmente gratuita.} (The use of this service is totally free of charge), the complement \textit{de este servicio} (of this service) corresponds to the original verb's direct complement. Similarly, in \textit{la duración de la estancia prevista} (the duration of the intended stay), the complement \textit{de la estancia prevista} (of the intended stay) corresponds to the subject of the verb \textit{durar} (to last). For this reason, to better direct users, the recommendation could emphasize nominalizations with such complements.
    
    Additionally, the exclusion list could include the term \textit{prestación} (allowance) and its plural, as they are frequently used in administrative contexts and lack a verb that can satisfactorily enhance clarity.
\end{itemize}

In considering the additional recommendations for simplifying the LengClaro versions, some of them are challenging to implement in a tool like arText claro. For instance, at the discourse level, it is particularly challenging to discern when a text fails to "get to the point" or when it exhibits inconsistencies resulting from "copy and paste" practices. Conversely, other recommendations may be more feasible to implement at different linguistic levels. Specifically, at the morphosyntactic level, the following suggestions could be addressed:

\begin{itemize}
    \item \textbf{Systematic use of second person verbs}. The system could include the detection of texts that combines verb forms of the 2nd person singular \textit{tú} with references to \textit{usted}, similar to how it currently identifies verb forms corresponding to the 1st person plural and singular.
    \item \textbf{Avoid parenthetical remarks}. The system could identify information within parentheses and either suggest removing it or adapting it in accordance with recommendation b9, depending on whether the user considers it to be irrelevant or not.
\end{itemize}

At the lexical level, a recommendation could be incorporated to the writing assistant:
\begin{itemize}
    \item \textbf{Avoid using foreign words.} Based on issues identified in the analysis of recommendation c2, arText claro could benefit from incorporating a new lexical-level recommendation feature. This feature would detect the presence of foreign words in a text and advise the user to avoid them. Given the predominance of English, a generic English lexicon could be employed for this purpose.
\end{itemize}

In addition, we observed writing errors in the analysed texts that contribute to the lack of clarity. These errors include typos, misspellings, grammatical mismatches, and poor punctuation. For the sake of clarity, it is important to avoid such mistakes. While some of these errors are detected by the arText claro text editor, not all of them are identified.
Most of the cases encountered are typos or problems with accents, as illustrated in the following example: \textit{Mas información en: [link]}. In this sentence, it should be written \textit{más}, a comparative adjective, but instead we find \textit{mas}, an adversative conjunction. The arText claro system incorporates a spell checker but does not directly recommend that users give importance to its warnings. It may be beneficial to explicitly recommend paying attention to this issue.

Grammatical mismatches are a source of inconsistencies that affect textual cohesion and coherence. If, at the sentence level, the words do not relate logically, it will be difficult for the text to function at a higher level of linguistic complexity. Most text editors nowadays include a grammar checking section that detects these types of mismatches in certain constructions. However, arText claro cannot make recommendations to correct errors in grammatical agreement. Below, we present some examples found. The fragments with an error are in bold:
    
    \begin{quote}
    \begin{itemize}
        \item Además, las víctimas de violencia contra la mujer pueden solicitar, a través de nuestras Oficinas de la Seguridad Social, que \textbf{el acceso a sus datos queden especialmente limitado y controlado}.

        \item Finalizado el proceso de cumplimentación \textbf{se le indicará los documentos} que debe presentar y tendrá la posibilidad de adjuntarlos electrónicamente.
    \end{itemize}
    \end{quote}
    
Proper punctuation is crucial for correctly delimiting ideas. For a text to be clear, it is as essential as using a simple lexicon. Some cases of ambiguity in the analyzed texts arise from difficulties in determining where one concept ends and the next begins. Below is an example where a clarification (in bold) is not correctly delimited by commas:

    \begin{quote}
    En todo caso, se procederá a la suspensión cautelar en el caso de traslado al extranjero por un periodo \textbf{continuado o no} superior a 90 días naturales al año, sin haber comunicado a la entidad gestora con antelación el mismo ni estar debidamente justificado.
    \end{quote}
    In this example, the absence of commas creates ambiguity in the interpretation of the disjunction. The most straightforward interpretation is:

    \underline{Interpretation (a)}. The disjunction separates the cases as follows:
    \begin{itemize}
        \item \textit{un periodo continuado} (a continuous period)
    
        or
        \item \textit{un periodo no superior a 90 días naturales} (a period not over 90 calendar days)
    \end{itemize}

    With a deeper understanding of the context, one may find that an alternative interpretation is more logical, even though the necessary commas for this interpretation are absent:
    
    \underline{Interpretation (b)}. The disjunction separates the cases as follows:
    \begin{itemize}
        \item \textit{un periodo continuado superior a 90 días naturales} (a continuous period over 90 calendar days)

        or
        \item \textit{un periodo no continuado superior a 90 días naturales} (a non-continuous period over 90 calendar days)
    \end{itemize}
    
    In fact, the translation to English en the Social Security website supports this second interpretation. In this case the remark is between commas (in bold):
    \begin{quote}
        In any case, the benefit shall be provisionally suspended in the event of transfer abroad for a period over 90 calendar days\textbf{, whether continuous or not,} without having notified the managing body in advance and without due justification.
    \end{quote}    

\subsection{Linguistic challenges}
\label{sec:linguistic challenges}

The primary challenge in these simplification processes was resolving ambiguities present in the original texts, particularly since those responsible for the simplifications are not the original authors of the texts. In some cases, ambiguities could be resolved through the application of common sense. In other instances, we needed to consult a legal expert to determine the precise scope of certain terms. Another strategy involved referring to versions of the same text available in other languages, such as English or French, to observe how the ambiguity was addressed in those cases. However, this approach carries certain risks. It is likely that these texts are translations based on the same Spanish source in which we identified the ambiguity. Without knowing the origin of the translation, we cannot determine whether it was produced through automatic means, whether the translator was someone within the administration familiar with the nuances of the content, or whether they faced the same challenges as we did. Consequently, we remain uncertain about the methods used to resolve the ambiguity.

Where possible, external sources were used to resolve ambiguities. In some cases, referring to the official state gazette (BOE) provided a resolution. For example, consider the phrase:
\begin{quote}
    Mujeres mayores de edad víctimas de violencia de género o víctimas de trata de seres humanos y explotación sexual.
\end{quote}

There are several possible interpretations:

\begin{enumerate}
    \item Women of legal age who are victims of either gender-based violence or human trafficking and sexual exploitation.
    \item Women of legal age who are victims of gender-based violence, and individuals (regardless of age or gender) who are victims of human trafficking and sexual exploitation.
\end{enumerate}

According to the BOE\footnote{\href{https://www.boe.es/diario\_boe/txt.php?id=BOE-A-2021-21007}{https://www.boe.es/diario\_boe/txt.php?id=BOE-A-2021-21007}}: "The age requirement will not apply [...] in the case of women who are victims of gender-based violence or human trafficking and sexual exploitation." Therefore, we conclude that the correct interpretation is option 1. To eliminate this ambiguity, we simplified the text as follows:
\begin{quote}
    Una mujer mayor de 18 años víctima de violencia de género o de trata de seres humanos y explotación sexual.
\end{quote}

In other cases, consulting the official gazette alone was not sufficient. For instance, in the original Document 1, there are clues that suggest that the child benefit supplement can be claimed independently of the MVI, although this possibility is not explicitly stated in the text. Some of these clues are linguistic, particularly pragmatic. The text mentions the incompatibility of the economic allocation per child or minor under charge with the MVI, and then reiterates the incompatibility of this allocation with the child benefit supplement:
\begin{quote}
    \begin{itemize}
        \item La percepción de la prestación de ingreso mínimo vital será incompatible con la percepción de la asignación económica por hijo o menor a cargo, sin discapacidad o con discapacidad inferior al 33 por ciento, cuando exista identidad de causantes o beneficiarios de esta.
        \item El complemento de ayuda para la infancia será incompatible con la asignación económica por hijo o menor a cargo sin discapacidad o con discapacidad inferior al 33 por ciento.
    \end{itemize}
\end{quote}

This is a conventional implicature: by separately mentioning their incompatibility with the economic allocation per child or minor under charge, it can be inferred that the MVI and the child benefit supplement are distinct benefits. Additionally, the text indicates that the Social Security service can be accessed to claim one "and/or" the other. This possible disjunction again implies that the child benefit supplement is not necessarily linked to the MVI:

\begin{quote}
    Para solicitar el ingreso mínimo vital \textbf{y/o} el complemento de ayuda para la infancia acceda al Servicio Ingreso Mínimo Vital en nuestra sede electrónica.
\end{quote}

Other clues in the text suggesting that the child benefit supplement is not dependent of the MVI are based on logical reasoning rather than linguistic analysis. For example, the income and asset requirements for qualifying for the child benefit supplement are less stringent than those for the MVI. This implies that a household unit could be eligible for the former without meeting the criteria for the latter\footnote{Income requirements for MVI: \textit{[...] el promedio mensual de ingresos y rentas anuales computables del ejercicio anterior sea inferior al menos en 10 euros a la cuantía mensual garantizada por el ingreso mínimo vital [...].}

Income requirements for the child benefit supplement: \textit{[...] en el ejercicio inmediatamente anterior al de la solicitud los ingresos computables sean inferiores al 300\% de la cuantía garantizada por el ingreso mínimo vital [...].}

Assets requirements for MVI: \textit{[...] no se considera en situación de vulnerabilidad económica cuando sean titulares de un patrimonio, sin incluir la vivienda habitual, valorado en un importe igual o superior al que se indica en esta tabla [...].}

Assets requirements for the child benefit supplement: \textit{[...] el patrimonio neto sea inferior al 150\% de los límites señalados anteriormente [...].}}. If the child benefit supplement were intrinsically linked to the MVI, it would be illogical to have more permissive requirements for it.

To ensure that these clues were correct and not redundant or outdated information, as observed elsewhere in the text, we needed to consult external sources. In this instance, confirmation was found in "La Revista de la Seguridad Social"\footnote{\href{https://revista.seg-social.es/-/gu\%C3\%ADa-sobre-el-nuevo-complemento-a-la-infancia-del-imv}{https://revista.seg-social.es/-/gu\%C3\%ADa-sobre-el-nuevo-complemento-a-la-infancia-del-imv} [Accessed: 17th June 2024]}: \textit{El complemento de Ayuda para la Infancia forma parte del Ingreso Mínimo Vital (IMV), aunque se puede conceder de forma independiente.} (The child benefit supplement is part of the Minimum Vital Income (MVI), although it can be granted independently.)

As demonstrated in this example, presenting information clearly is crucial to ensure that it is effectively communicated to the audience. Using terms such as "supplement" to refer to a separate benefit, or employing very similar terminology for different, incompatible child benefits, complicates this goal. Even if the information in a text is accurate, it may not reach the audience if not presented clearly. 

%Por último, ha ha habido casos en los que no nos ha sido posible resolver la ambigüedad para evitarla de las versiones simplificadas.

Another challenge arising from simplifying texts without being the original authors is that there have been instances where we lacked the necessary administrative background knowledge to accurately assess the relevance of certain text sections. At times, our judgment of what may be considered superfluous—and thus a candidate for removal in a simplified version—might overlook specific rules or nuances that administrative staff would be aware of. Conversely, there may be instances where we have retained content in the text because we deemed it important, even though it could have been omitted.

\section{Conclusion and Future Work}
\label{sec:conc}

In this work we have presented the development of LengClaro2023, a dataset composed of administrative texts in Spanish and their corresponding simplified versions, aimed at enhancing clarity and accessibility for the general public. LengClaro3023 includes two distinct simplified versions: one generated through arText claro, a tool that automatically  detects features that hinder clarity and suggests corrections, and the other produced by manually applying a set of plain language recommendations for legal-administrative texts.

Qualitative analysis of the simplification process revealed that, in addition to the typical characteristics of legal-administrative language that generate uncertainty for readers, many issues of obscurity in the analyzed texts stem from poor use of written language, such as incorrect verb tenses, inaccurate punctuation, and inconsistencies in concordance.

Furthermore, digital texts exhibit specific characteristics that distinguish them from traditional paper documents. The manner in which audiences engage with this type of text—primarily through screen reading—affects how they comprehend the content. This change, coupled with the unique capabilities of websites, necessitates the use of particular strategies to enhance clarity in digital texts.

The analysis of the writing assistant arText claro from a user's perspective revealed several limitations and challenges:

\begin{itemize}
    \item arText claro can process long texts but struggles with very lengthy documents and is limited in the formats it can handle, as it does not support HTML.
    \item The tool suffers from technical issues, the origins of which are difficult to identify, impacting its overall reliability.
    \item It has difficulty accurately identifying sentence and paragraph boundaries and shows inconsistency in recognizing features such as acronyms or complex passive constructions.
    \item Some of arText claro's recommendations are challenging for users to adhere to due to their strictness, such as in the case of sentence length, or because they are too generic, lacking precise flagging of problematic elements, such as participles.
    \item The most significant limitation observed in using arText claro is its inability to provide recommendations on text organization, which was a major issue in the original texts within our dataset.
\end{itemize}

However, it is important to note that the manual simplification process for generating LengClaro versions was significantly time-consuming, with an estimated duration ranging from 30 minutes to an hour and a half for content equivalent to one sheet of paper. This highlights a significant trade-off between the quality of simplification and the time investment required. For this reason, arText claro, despite its limitations, remains a valuable resource, as it serves as a helpful starting point for simplification, reducing the overall effort required for manual editing and refinement.

In conclusion, this work contributes to the field of text simplification by not only creating accessible content for the citizens but also by providing a qualitative analysis on plain language recommendations and linguistic challenges in text simplification that informs future tools and methodologies. In addition, the resulting dataset, structured in sets of HTML trios, enables the pairing of these trios as complex-simple binomials (e.g., original-artext or original-LengClaro). This structure makes this dataset a valuable resource for evaluating and improving machine learning based automatic text simplification (ATS) systems \citep{gonz-del-nav-col2024} aimed at making complex administrative language more understandable to a broader audience. The findings underscore the importance of human involvement in the simplification process while also acknowledging the practical value of NLP tools like arText claro.

Beyond these contributions, there is significant scope for future research and development. Currently, the size of LengClaro2023 is quite limited due to the scope of this work. Expanding it to include a wider range of administrative texts in Spanish, along with their corresponding simplified versions, would be beneficial. In addition to the document-level dataset, a manually aligned sentence-level dataset could be developed. These enhancements would transform the dataset into a more powerful resource for training and testing ATS systems.

In addition, according to the preliminary tests with recent LLMs conducted, there is still work to do to enhance the performance of tools like ChatGPT and Phi-2-LC in supporting text simplification tasks. Approaching these improvements from a prompt engineering perspective appears promising.

Regarding manuals for Plain Language in Spanish, it is important to highlight that in February 2024, the Spanish Standardisation Agency (UNE) published the Spanish version of the ISO Plain Language Standard, originally released in June 2023: UNE-ISO 24495-1:2024 Plain Language—Part 1: Governing Principles and Guidelines\footnote{\href{https://www.une.org/encuentra-tu-norma/busca-tu-norma/norma?c=N0072523}{https://www.une.org/encuentra-tu-norma/busca-tu-norma/norma?c=N0072523}}. The revision conducted in this work did not include it due to its recent release. Similarly, the \textit{Guía panhispánica de lenguaje claro y accesible} \citep{guiaPanhispanica2024}, which aims to "explain plain language using plain language" and is scheduled for publication on 25th September 2024. Reviewing these sources could offer valuable insights for future improvements in plain language usage within Spanish administrative contexts.

By exploring these future directions, researchers can continue to advance the field of text simplification, making important documents more accessible and comprehensible to a wider audience. This can help empower individuals and promote greater transparency.

%\clearpage
\section*{Limitations}

This report presents a dataset of texts from the Spanish Government adapted to plain language in peninsular standard Spanish. The rules and criteria used to simplify the texts do not follow the  UNE-ISO 24495-1:2024 Plain Language—Part 1: Governing Principles and Guidelines, since this UNE rule was published after the dataset was created. The texts have been simplified by a professional translator with a background in Plain Language. The texts included in the data set may have been updated in their original website and legal information may have been modified.

%\section*{Short data statement}

%\textcolor{red}{XXXX}

%\section*{Acknowledgments}

%\textcolor{red}{XXXX}

% Bibliography entries for the entire Anthology, followed by custom entries
%\bibliography{anthology,custom}
% Custom bibliography entries only
\bibliography{custom}

%\appendix

%\section{Example Appendix}
%\label{sec:appendix}

%This is an appendix.

\end{document}